\documentclass[letterpaper, 10 pt, conference]{ieeeconf}  

\IEEEoverridecommandlockouts                              

\usepackage{graphicx}
\usepackage{caption}
\usepackage{bm}
\usepackage{amsmath, amssymb, amsfonts}

\usepackage{tabularx}
\usepackage{multirow, multicol}
\usepackage{adjustbox}

\usepackage[linesnumbered,ruled,vlined]{algorithm2e}
\usepackage{setspace}
\usepackage[T1]{fontenc}

\usepackage[maxfloats=256]{morefloats}
\usepackage{dsfont}

\usepackage{subfigure}
\usepackage{multirow}
\usepackage{hyperref}

\title{\LARGE \bf
Design and Evaluation of an Uncertainty‑Aware Shared‑Autonomy System with Hierarchical Conservative Skill Inference
}

\author{Taewoo Kim, Donghyung Kim, Minsu Jang and Jaehong Kim
\thanks{Taewoo Kim, as a Senior Researcher, is with the Social Robotics Research Section, Electronics and Telecommunications Research Institute (ETRI), Daejeon, Republic of Korea (e-mail: {\tt\small twkim0812@etri.re.kr}; {\tt\small twkim0812@gmail.com})}%
\thanks{Donghyung Kim, as a Senior Researcher, is with the Field Robotics Research Section, Electronics and Telecommunications Research Institute (ETRI), Daejeon, Republic of Korea (e-mail: {\tt\small donghyungkim@etri.re.kr})}%
\thanks{Minsu Jang, as a Principal Researcher, is with the Social Robotics Research Section, Electronics and Telecommunications Research Institute (ETRI), Daejeon, Republic of Korea (e-mail: {\tt\small minsu@etri.re.kr})}%
\thanks{Jaehong Kim, as a Principal Researcher and Director, is with the Social Robotics Research Section, Electronics and Telecommunications Research Institute (ETRI), Daejeon, Republic of Korea (e-mail: {\tt\small jhkim504@etri.re.kr})}%
}

\begin{document}

\maketitle
\thispagestyle{empty}
\pagestyle{empty}

\begin{abstract}

Shared‑autonomy imitation learning lets a human correct a robot in real time, mitigating covariate‑shift errors. Yet existing approaches ignore two critical factors: (i) the operator’s cognitive load and (ii) the risk created by delayed or erroneous interventions. We present an uncertainty‑aware shared‑autonomy system in which the robot modulates its behaviour according to a learned estimate of latent‑space skill uncertainty. A hierarchical policy first infers a conservative skill embedding and then decodes it into low‑level actions, enabling rapid task execution while automatically slowing down when uncertainty is high. We detail a full, open‑source VR‑teleoperation pipeline that is compatible with multi‑configuration manipulators such as UR‑series arms. Experiments on pouring and pick‑and‑place tasks demonstrate 70–90\% success in dynamic scenes with moving targets, and a qualitative study shows a marked reduction in collision events compared with a non‑conservative baseline. Although a dedicated ablation that isolates uncertainty is impractical on hardware for safety and cost reasons, the reported gains in stability and operator workload already validate the design and motivate future large‑scale studies. 

\noindent\textbf{Resources} – : \href{https://github.com/aai4r/aai4r-pouring-skill}{Github Code}, \href{https://www.youtube.com/watch?v=2Gwvn6xzEZ0}{Demo video}.



\end{abstract}

\section{INTRODUCTION}

Imitation learning has been widely applied across various domains such as autonomous vehicles \cite{qureshi2023imitation} and robotic tasks \cite{wang2022adaptive, stepputtis2020language} as an effective method for learning a target task with guidance from experts. Recently, a shared autonomy process (SAP) \cite{jang2022bc} has been proposed, based on the imitation learning and the HG-DAGGER \cite{kelly2019hg}, where experts and robotic agents share the workspace and experts correct motions. Expanding on this, language-based manipulation skills have been developed through the large-scale demonstration dataset \cite{jang2022bc}. In this manner, recent research trends, such as large language models (LLMs) \cite{zhao2023survey}, demand extensive datasets for training more generalized artificial intelligence. However, in the field of robotics, the diversity of experimental environments has led to a scarcity of publicly available datasets that can be universally utilized. While there are methods to collect datasets through simulation environments, setting up such environments can be time-consuming and costly. Moreover, the domain gap issues between simulations and real-world still exists. Hence, the process of collecting datasets through interaction with real-world robots is inevitable.

\begin{figure}
	\begin{center}
		\includegraphics[width=\linewidth]{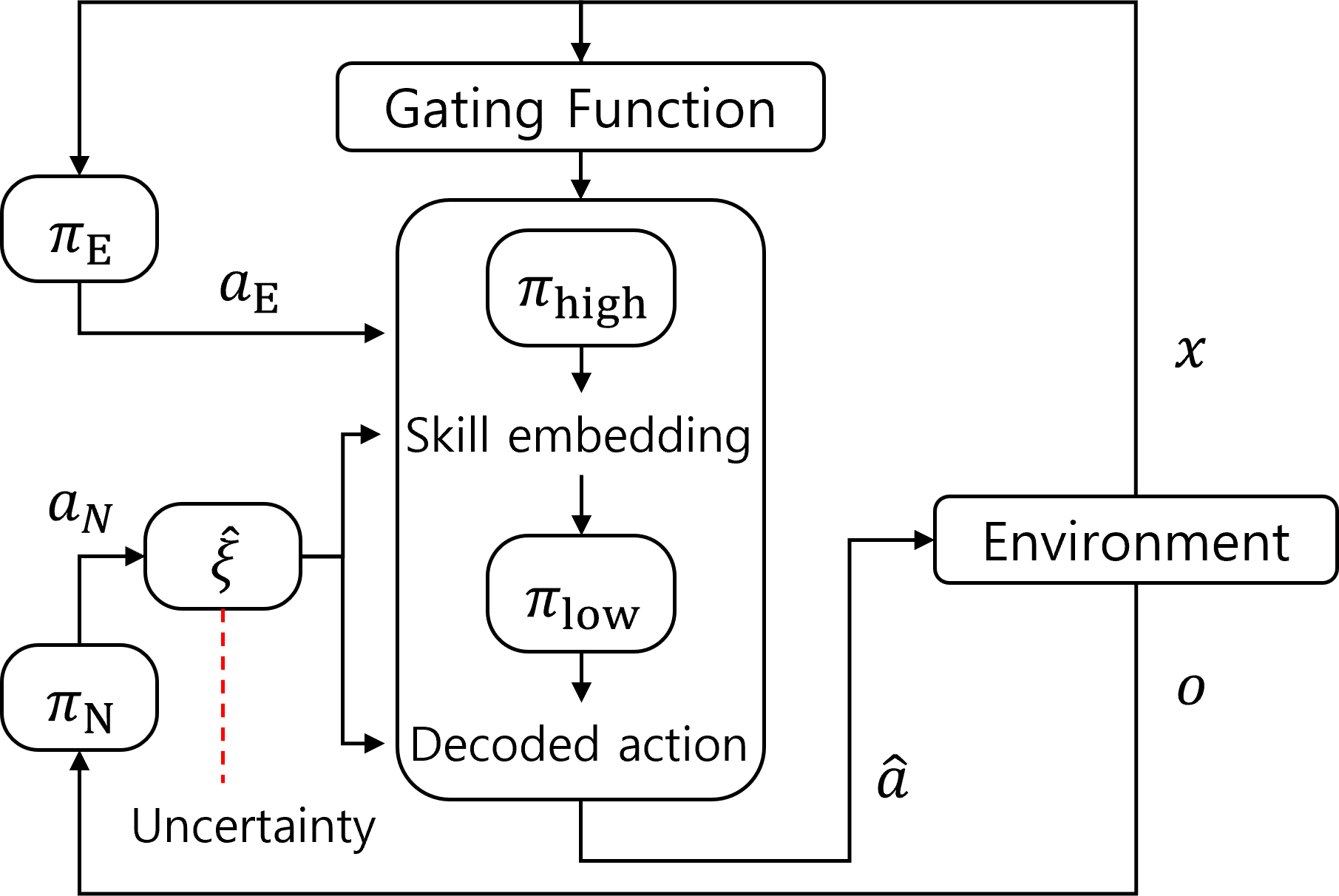}
	\end{center}
	
	\caption{Control loop for a Hierarchical Conservative Skill Network in Uncertainty-Aware Shared Autonomy Process}
        \label{fig:conf_unc_skill}
	\vspace{-1.5em}
\end{figure}

 In state-of-the-art researches, \cite{brohan2022rt, brohan2023rt}, they showed outstanding results in learning robotic manipulation skills through the large-scale demonstrations gathered from the SAP framework. SAP effectively improves the policy actions in unvisited states and the associated compounding errors by allowing user intervention and control acquisition in the event of anticipated robot task failures or hazardous situations during skill demonstrations, followed by motion corrections (Fig. \ref{fig:conf_unc_skill}) \cite{kelly2019hg, jang2022bc}. However, this approach requires prolonged interactions with the robot and continuous supervision, which not only leads to high levels of fatigue for the operator \cite{bradbury2016attention} but also does not account for the possibility of errors by the expert \cite{reason2000human}. Particularly, in the process of handling robots in real-world environments, it is imperative to consider safety issues. For instance, delayed decision-making regarding operator intervention can result in damage to the robot and the environment or cause injury to the operator. Furthermore, when applied in real-world settings, robots may encounter various dynamic environmental changes. Therefore, it is essential to consider issues related to robot control in uncertain situations. However, Recent SAP-based studies did not specifically address these concerns in the dataset collection and application process  \cite{jang2022bc, brohan2022rt, brohan2023rt}.

\begin{figure}
	\begin{center}
		\includegraphics[width=\linewidth]{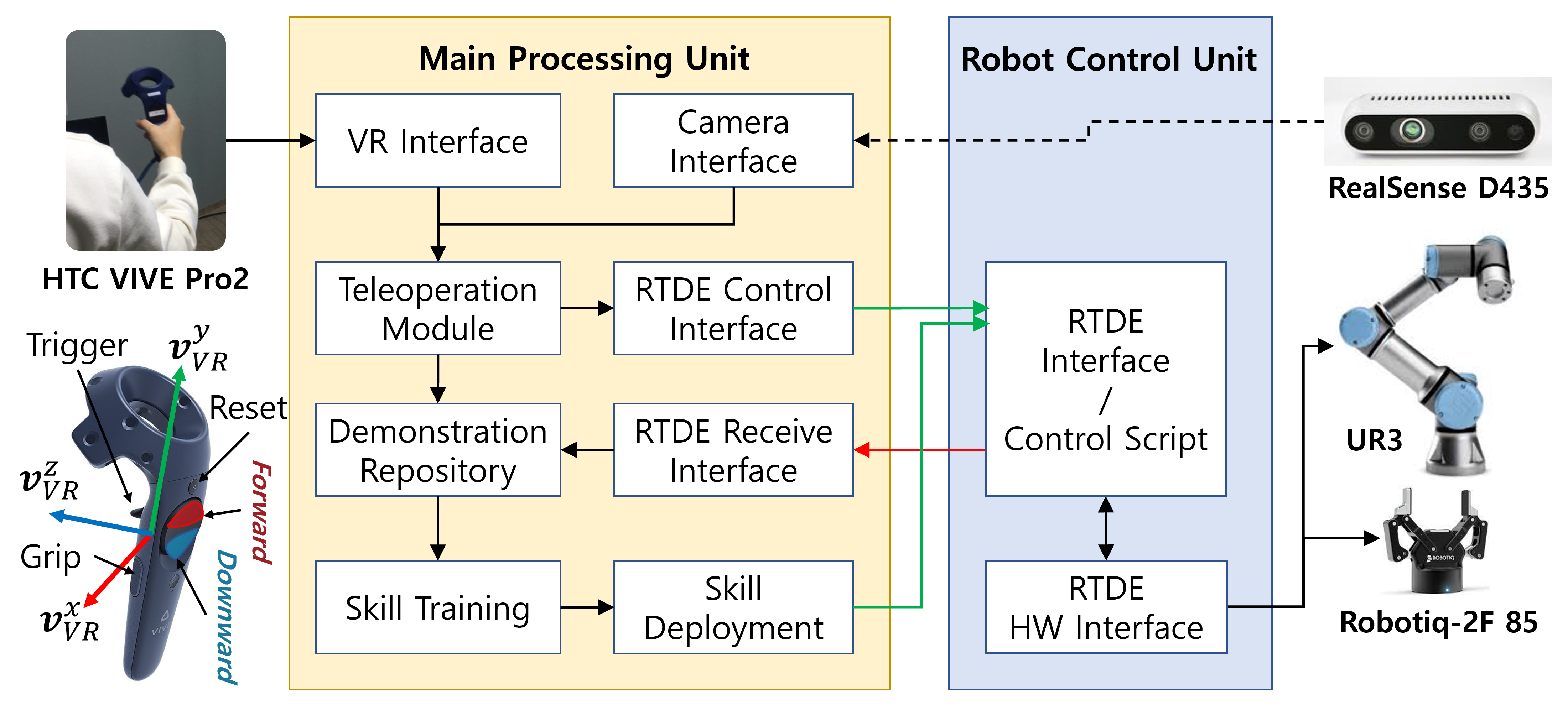}
	\end{center}
	\vspace{-0.5em}
	\caption{The comprehensive structure of our SAP system and the configuration of the VR controller.}
        \label{fig:architecture}
	\vspace{-1.5em}
\end{figure}

In this study, we propose an imitation learning approach that enables the robot agent to infer uncertainty and, consequently, perform manipulation skills more conservatively. This approach is aimed at addressing human errors and associated safety issues that may arise during the SAP-based learning process. Modeling robot skills and planning problems solely based on the end-effector trajectory for a specific manipulation task lacks systematicity and scalability. Therefore, we introduce a hierarchical network structure inspired by SPiRL \cite{pertsch2021accelerating} to facilitate learning and inference at the abstracted skill level of the robot. Our hierarchical skill network is divided into high-level and low-level policies. Each is responsible for generating abstracted skill embeddings from environmental input information and subsequently decoding them into actual robot behaviors.

To facilitate uncertainty inference, we applied Monte-Carlo dropout \cite{gal2016dropout} to the high-level policy network, which operates between the environmental state information and the skill embedding, due to its simplicity. Through this approach, we infer uncertainty at the skill level and design a more conservative planning strategy at the skill level based on the degree of uncertainty. Additionally, we apply conservative action inference, accounting for uncertainty, to the final robot action input (Fig. 1). Through the proposed hierarchical structure and conservative skill inference method, we experimentally demonstrated the stability of the learning process based on SAP. As a result, we propose an approach that can increase the tolerance for human errors.

To validate the proposed method, we constructed a SAP-based manipulator teaching system from scratch, utilizing virtual reality (VR) teleoperation. In the RT studies, the SAP system played a crucial role in conducting the study; however, it was described in only a few lines, and only the source code related to the learning was made public. In this paper, we provide a detailed description of the SAP system constructed directly using a VR device and Universal Robot (UR3), and we make all relevant source code publicly available. Furthermore, we outline detailed system designs that consider the multi-configuration modes, specifically both \textit{forward} and \textit{downward}, while taking into account the minimal required operation stability, with due consideration to the configuration of UR3.

    

\section{Shared Autonomy System}

\subsection{System Overview}
Our system consists of a primary processing unit (MPU) and a robot control unit (RCU), as depicted in Fig. \ref{fig:architecture}. The MPU comprises several components, including VR and camera interfaces, teleoperation, a demonstration repository, skill learning, and real-time data exchange (RTDE). The VR interface captures the user's motion with synchronized scene images from the camera interface and conveys this data to the teleoperation module. In the teleoperation module, user input motion is converted into robot motion commands, then sent to the RCU via the RTDE. After completing a task demonstration, all teleoperation data, including VR motion, images, and proprioceptive robot states (e.g., joint angles), is stored in the demonstration repository. The skill training module then uses these demonstrations to acquire manipulation skills. The RCU handles communication with the MPU through RTDE, processes user commands, and executes robot control.

\begin{figure}
	\begin{center}
		\includegraphics[width=\linewidth]{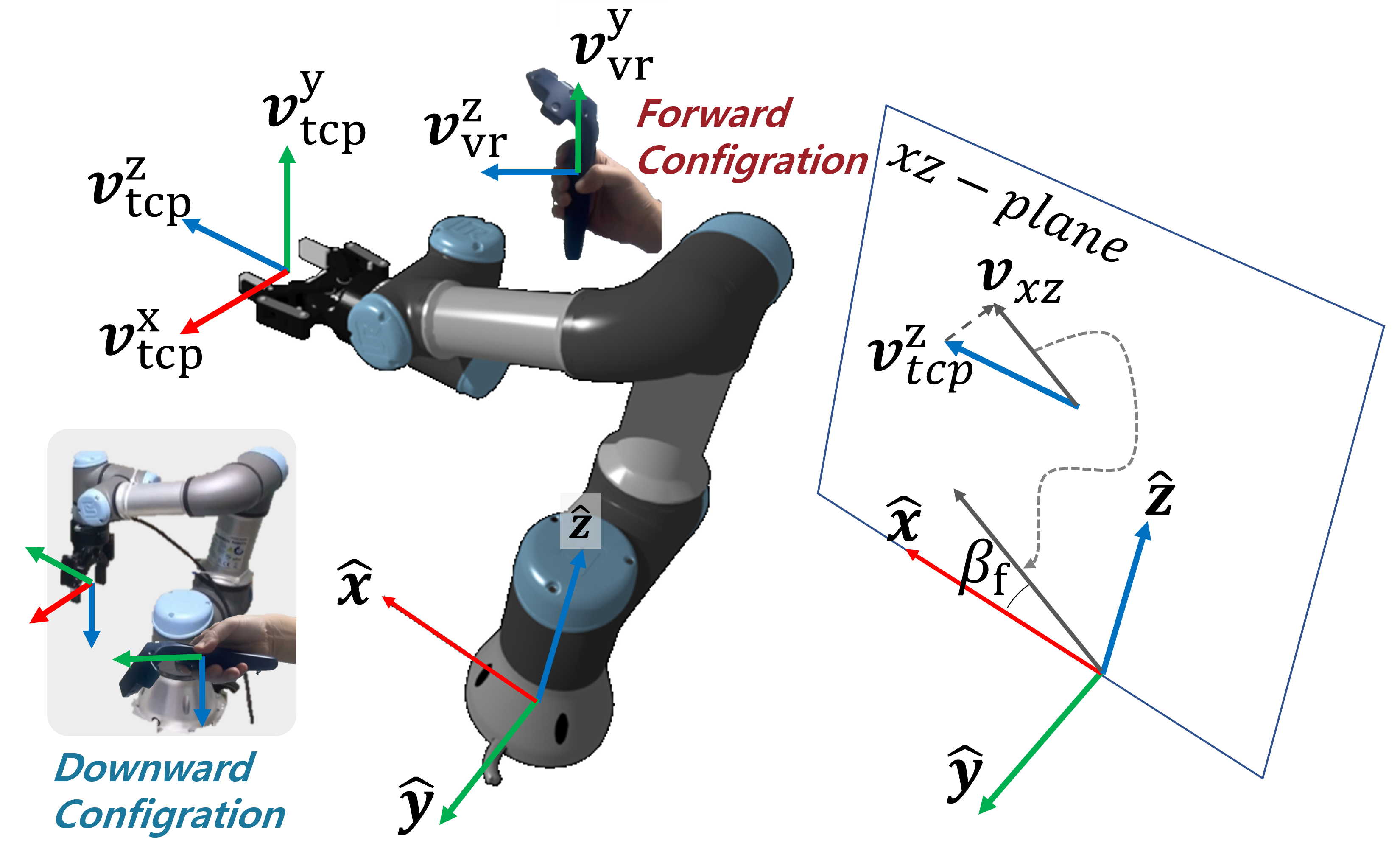}
	\end{center}
	\vspace{-0.5em}
	\caption{Visualization of beta constraint, forward and downward configurations, and corresponding VR controller base postures. }
        \label{fig:beta_const}
\end{figure}


\subsection{VR Teleoperation Interface}
In the MPU, the VR interface processes various user commands including controller motion and button events. We conducted teleoperated demonstrations utilizing only the controller without the head-mounted display (HMD), relying on direct observation of Human eyes. As shown in Fig. \ref{fig:architecture}, four dedicated buttons are used for robot teleoperation. The menu button resets the current demonstration episode, the trigger button gives the operator control of the slave robot, the grip button opens and closes the slave robot gripper, and the trackpad button switches the robot's configuration between forward and downward during teleoperation. Despite HTC VIVE's left-handed coordinate system, we converted it to a right-handed system to align the local controller coordinate system. 


\setlength{\textfloatsep}{5pt}
\begin{algorithm}
    \setstretch{0.9}
    \DontPrintSemicolon 
    Initialize $\mathcal{D}$, robot and demo count $N$\;
    \For{$n \gets 0$ \textbf{to} $N$} {
        $t \gets 0$ \;
        \While {\text{true}} {
            \If{$\mathrm{vr}_{\text{reset}} == \text{true}$}{
                $ \tau_T \gets \{\boldsymbol{o}_{T}, \boldsymbol{s}_{T}, \boldsymbol{a}^{\text{null}}_{T}, u^{\text{true}}_T$ \}\;
                $\mathcal{T}_n.\mathrm{insert}(\boldsymbol{\tau})$ \;
                $\mathcal{D}.\mathrm{insert}(\mathcal{T}_n) \text{ and then } \mathcal{T}_n \gets null$\;
                $\boldsymbol{\theta} \gets \boldsymbol{\theta}_{\text{init}} + \epsilon{;} \quad \Dot{\boldsymbol{\theta}} \gets 0$ \;
                \textbf{break}
            }
            $\Dot{\boldsymbol{\theta}} \gets 0, \: \boldsymbol{a}_t \gets null$\;
            \If{$\mathrm{vr}_{\mathrm{trigger}} == \text{true}$}{
                $a^{\text{mode}}_t \gets \mathrm{vr}_{\text{mode}}{;} \quad 
                    a^{\text{grip}}_t \gets \mathrm{vr}_{\text{grip}}$ \;
                \If{$a^{\mathrm{mode}}_t != \mathrm{mode}_t$}{
                    $\boldsymbol{\theta} \gets \text{ModeToJoint}(a^{\text{mode}}_t)$ \;
                    $\text{RCU}(\boldsymbol{\theta})$ \;
                }
                $\Dot{\boldsymbol{p}}^{\text{act}}_t \gets \Dot{\boldsymbol{p}}^{\mathrm{vr}}_t {;} \quad 
                    \boldsymbol{q}^{\text{\text{act}}}_t \gets \boldsymbol{q}^{\text{vr}}_t \otimes {\boldsymbol{q}^{\text{tcp}}_t}^{*}$ \;
                $\boldsymbol{a}^{\text{act}}_{t} = \{\Dot{\boldsymbol{p}}^{\text{act}}_t, \: \boldsymbol{q}^{\text{act}}_t, \: a^{\text{grip}}_t \:, a^{\text{mode}}_t \}$ \;
                $\tau_t \gets \{\boldsymbol{o}_{t}, \boldsymbol{s}_{t}, \boldsymbol{a}^{\text{act}}_{t}, u^{\text{false}}_t \} $\;
                
                $\boldsymbol{p}^{'}_{\mathrm{tcp}} \gets \boldsymbol{p}^{\mathrm{tcp}}_t + \boldsymbol{a}^{act}_t {;} \quad         
                    \boldsymbol{q}^{'}_{\mathrm{tcp}} \gets \boldsymbol{q}^{\text{vr}}_t$  \;
                $\boldsymbol{p}^{''}_{\mathrm{tcp}}, \: \boldsymbol{q}^{''}_{\mathrm{tcp}} \gets \delta(\boldsymbol{p}^{'}_{\mathrm{tcp}}, \: \boldsymbol{q}^{'}_{\mathrm{tcp}})$ \;
                $\boldsymbol{\theta}_{\text{goal}} \gets \text{IK}(\boldsymbol{p}^{''}_{\mathrm{tcp}}, \: 
                    \boldsymbol{q}^{''}_{\mathrm{tcp}})$ \;
                $\Dot{\boldsymbol{\theta}} \gets (\boldsymbol{\theta}_{\text{goal}} - \boldsymbol{\theta}^{\text{tcp}}_t) \times scale $\;
                $t \gets t + 1$ \;
            }
            RCU($\Dot{\boldsymbol{\theta}}$)
        }    
    }
    \caption{Task Demonstration Dataset Collection}
\end{algorithm}

\subsection{Constrained Teleoperation}
Conventional robot teleoperation methods inherently entail collision risks, as they directly transmit the master device's motion to the slave robot without adequate safeguards. To address this concern, we developed a constrained teleoperation method designed to proactively prevent self-collisions and collisions with the floor during teleoperation. This method involves the imposition of virtual motion constraints on the slave robot, which are applied to the desired tool center point (TCP) pose derived from the VR controller motion:
\begin{align}
    \boldsymbol{p}^{'}_{\text{tcp}} &= \boldsymbol{p}_{\text{tcp}} + \boldsymbol{\Dot{p}}_{\text{vr}} \\
    \boldsymbol{q}^{'}_{\text{tcp}} &= \boldsymbol{q}_{\text{vr}}
\end{align}
\noindent where the desired TCP position, denoted as $\boldsymbol{p}^{'}_{\text{tcp}}$, is determined by adding the linear velocity of the VR controller, expressed as $\boldsymbol{\Dot{p}}_{\text{vr}}=\{\Dot{x}_{\text{vr}}, \; \Dot{y}_{\text{vr}}, \; \Dot{z}_{\text{vr}}\}$, to the current TCP position, which is represented as $\boldsymbol{p}_{\text{tcp}}=\{x_{\text{tcp}}, \; y_{\text{tcp}}, \; z_{\text{tcp}}\}$. Similarly, the desired TCP orientation, denoted as $\boldsymbol{q}^{'}_{\text{tcp}}$, is simply defined to match the current orientation of the VR device itself, represented as $\boldsymbol{q}_{\text{vr}}=\{x_{\text{vr}}, \; y_{\text{vr}}, \; z_{\text{vr}}, \; w_{\text{vr}}\}$, utilizing quaternion notation in practical implementation. Essentially, this setup implies that the TCP's positional movement is directly proportional to the master device's positional speed, while rotational adjustments remain in perfect synchronization with the master device. The positional constraints are calculated using a straightforward delta function:
\begin{align}
    \delta(x, \lambda_{\text{min}}, \lambda_{\text{max}}) &= \max(\min({x, \, \lambda_{\text{max}}}), \, \lambda_{\text{min}}) \\
      \Bar{p}^x_{\text{tcp}} &= \delta(p^x_{\text{tcp}}, \lambda^{x}_{\text{min}}, \lambda^{x}_{\text{max}})
\end{align}
\noindent where $p^x_{\text{tcp}} \in \boldsymbol{p}_{\text{tcp}}=\{p^x_{\text{tcp}}, p^y_{\text{tcp}}, p^z_{\text{tcp}}\}$ represents the scalar component of the desired TCP position. Additionally, we have $\lambda^{x}_{\text{max}} \in \boldsymbol{\lambda}_{\text{max}}=\{\lambda^{x}_{\text{max}}, \lambda^{y}_{\text{max}}, \lambda^{z}_{\text{max}}\}$ and $\lambda^{x}_{\text{min}} \in \boldsymbol{\lambda}_{\text{min}}=\{\lambda^{x}_{\text{min}}, \lambda^{y}_{\text{min}}, \lambda^{z}_{\text{min}}\}$, which correspond to the maximum and minimum thresholds, respectively.

While defining position constraints is straightforward, establishing rotation constraints demands a more intricate approach to ensure precise motion limitations. Specifically, for rotation constraints, we introduced the concept of plane-projected rotational constraints. To implement this, we began by defining projection matrices for three fundamental planes of the $xy$ plane, the $xz$ plane, and the $yz$ plane (Fig. \ref{fig:beta_const}):
\begin{equation} \label{eq2}
P_{xy} = (\hat{P}_{xy} (\hat{P}_{xy}^\intercal \hat{P}_{xy})^{-1}) \hat{P}_{xy}^\intercal
\end{equation}

The formulation of the rotation constraint involves the use of projection matrices for three base planes, including $\hat{P}_{xy}$, which is a $(3 \times 2)$ matrix containing the unit column vectors $\hat{\boldsymbol{x}}$ and $\hat{\boldsymbol{y}}$. Similar projection matrices for the $xz$ and $yz$ planes are established in a similar manner.

To implement the rotation constraint, we projected the TCP's coordinates onto these base planes using the respective projection matrices. This projection allowed us to compute the numerical angles corresponding to roll, pitch, and yaw, effectively constraining rotation. This process exhibits slight variations for each configuration mode, as outlined in equations (\ref{eq:beta_proj}) to (\ref{eq:alpha_dwn}).

Forward Configuration:
\begin{align}
    \boldsymbol{v}_{xz} &= P_{xz} \boldsymbol{v}^z_{\text{tcp}} \label{eq:beta_proj} \\
    \beta_f &= \arccos(\boldsymbol{v}_{xz} \cdot \hat{\boldsymbol{x}} / (||\boldsymbol{v}_{xz}|| \cdot ||\hat{\boldsymbol{x}}||)) \label{eq:beta_angle} \\
    \beta_f &= \delta(\beta_f \times \mathcal{C}(v^z_{xz} < 0), \lambda^{\beta_f}_{\text{min}}, \lambda^{\beta_f}_{\text{max}}) \label{eq:beta_clip} \\
    \boldsymbol{v}_{xy} &= P_{xy} \boldsymbol{v}^z_{\text{tcp}} \\
    \gamma_f &= \arccos(\boldsymbol{v}_{xy} \cdot \hat{\boldsymbol{x}} / (||\boldsymbol{v}_{xy}|| \cdot ||\hat{\boldsymbol{x}}||)) \\
    \gamma_f &= \delta(\gamma_f \times \mathcal{C}(v^y_{xy} > 0), \lambda^{\gamma_f}_{\text{min}}, \lambda^{\gamma_f}_{\text{max}}) \\
    \boldsymbol{v}_{yz} &= P_{yz} \boldsymbol{v}^y_{\text{tcp}} \\
    \alpha_f &= \arccos(\boldsymbol{v}_{yz} \cdot \hat{\boldsymbol{z}} / (||\boldsymbol{v}_{yz}|| \cdot ||\hat{\boldsymbol{z}}||)) \\
    \alpha_f &= \delta(\alpha_f \times \mathcal{C}(v^y_{yz} > 0), \lambda^{\alpha_f}_{\text{min}}, \lambda^{\alpha_f}_{\text{max}})
\end{align}
Downward Configuration:
\begin{align}
    \boldsymbol{v}_{xz} &= P_{xz} \boldsymbol{v}^y_{\text{tcp}} \\
    \beta_d &= \arccos(\boldsymbol{v}_{xz} \cdot \hat{\boldsymbol{x}} / (||\boldsymbol{v}_{xz}|| \cdot ||\hat{\boldsymbol{x}}||)) \\
    \beta_d &= \delta(\beta_d \times \mathcal{C}(v^z_{xz} < 0), \lambda^{\beta_d}_{\text{min}}, \lambda^{\beta_d}_{\text{max}}) \\
    \boldsymbol{v}_{xy} &= P_{xy} \boldsymbol{v}^y_{\text{tcp}} \\
    \gamma_d &= \arccos(\boldsymbol{v}_{xy} \cdot \hat{\boldsymbol{x}} / (||\boldsymbol{v}_{xy}|| \cdot ||\hat{\boldsymbol{x}}||)) \\
    \gamma_d &= \delta(\gamma_d \times \mathcal{C}(v^y_{xy} > 0), \lambda^{\gamma_d}_{\text{min}}, \lambda^{\gamma_d}_{\text{max}}) \\
    \boldsymbol{v}_{yz} &= P_{yz} \boldsymbol{v}^z_{\text{tcp}} \\
    \alpha_d &= \arccos(\boldsymbol{v}_{yz} \cdot -\hat{\boldsymbol{z}} / (||\boldsymbol{v}_{yz}|| \cdot ||\hat{\boldsymbol{z}}||)) \\
    \alpha_d &= \delta(\alpha_d \times \mathcal{C}(v^y_{yz} < 0), \lambda^{\alpha_d}_{\text{min}}, \lambda^{\alpha_d}_{\text{max}})
    \label{eq:alpha_dwn}
\end{align}

\begin{figure*}
	\begin{center}
		\includegraphics[width=1.0\linewidth]{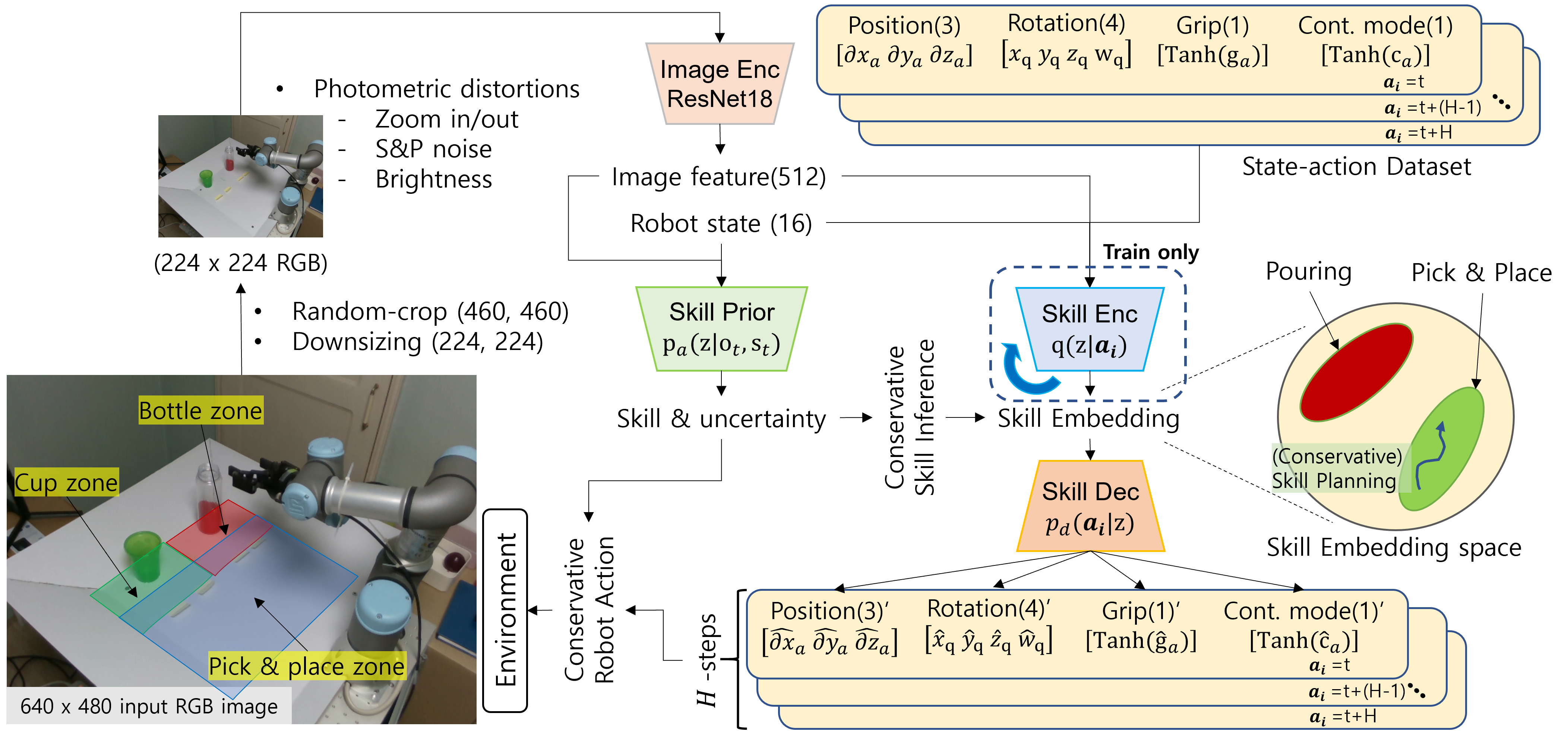}
	\end{center}
	\caption{Overall architecture of the hierarchical skill network (HSN).}
        \label{fig:skill_net}
        \vspace{-1.0em}
\end{figure*}

Subsequently, the rotation constraint is finalized by applying rotation clipping thresholds to the actual angle values, which are obtained from the projected TCP coordinates. For instance, in the case of $\beta_f$, the process involves projecting the TCP's $z$-axis direction vector $\boldsymbol{v}^z_{\text{tcp}}$ onto the $xz$-plane (Eq. \eqref{eq:beta_proj}). The angle value is then calculated through the dot-product of the projected vector $\boldsymbol{v}_{xz}$ and the basis vector $\hat{\boldsymbol{x}}$, followed by the arccos function (Eq. \eqref{eq:beta_angle}). The sign of this angle is set to positive when $v^z_{xz}$ is less than zero and negative otherwise by the conditional sign function $\mathcal{C}(x)=\left[(\mathds{1}|x=\text{T}) \: \text{or} \: (-\mathds{1}|x=\text{F}) \right]$, where $ v^z_{xz} \in \boldsymbol{v}_{xz}$. Finally, $\beta_f$ is confined within the range defined by $\lambda^{\beta_f}_{\text{min}}$ and $\lambda^{\beta_f}_{\text{max}}$ (Eq. \eqref{eq:beta_clip}). Additional details regarding the actual threshold values for pose constraints in the forward and downward configurations are presented in Table \ref{table:const_min_max_thres_all}.

\subsection{Human Demonstration Dataset Collection}
Human demonstrations for initial skill policy learning were recorded at 30Hz and saved to disk at the end of each episode. Each episode includes TCP trajectories, consisting of observation, robot state, action, and an episode end flag. This data forms the dataset for the target task, denoted as a rollout memory $\mathcal{D}=\{\mathcal{T}_1, \dots, \mathcal{T}_N\}$, where each trajectory $\mathcal{T}_n$ comprises packets $\tau_t=\{o_t, ; s_t, ; a_t, u_t\}$. A packet includes an RGB image for observation, a 16-dimensional state vector (joint angles, TCP position with quaternion orientation, normalized gripper position, and configuration mode), a 9-dimensional action vector (position, rotation, gripper, and conf. mode), and an episode end flag. The rotation action follows the VR controller's orientation, with the recorded rotation action determined by the difference between TCP and VR controller orientations: 
\begin{align}
    \boldsymbol{q}^{\text{vr}}_t &= \boldsymbol{q}^{\text{act}}_t \otimes \boldsymbol{q}^{\text{tcp}}_t
\end{align}
where the unknown rotation action $\boldsymbol{q}^{\text{act}}_t$ is calculated by multiplying the conjugate of the TCP orientation $\boldsymbol{q}^{\text{tcp*}}_t$ on both sides of the equation. Algorithm 1 outlines the demonstration dataset collection process.

\section{Hierarchical Conservative Skill Inference}
\subsection{Hierarchical Skill Network Framework}
To learn the manipulation skills from human demonstrations, we designed a hierarchical skill network (HSN) model (Fig. \ref{fig:skill_net}) inspired from the SPiRL architecture \cite{pertsch2021accelerating}. HSN consists of a hierarchical structure that infers robot skill embedding from observations of the environment and decodes it into actual robot actions, thereby controlling the real robot. In training phase, HSN learns skill embedding space (pouring and pick and place skills in our case) using recurrent skill encoder $q(z|\bold{a_i})$ and skill decoder $p_d(\bold{a_i}|z)$ while the skill prior $p_a(z_t|o_t, s_t)$ is trained to learn skill embedding distributions corresponding to the observations by minimizing the Kullback-Leibler divergence \cite{csiszar1975divergence} between the predicted prior and the inferred skill posterior $\mathbb{E}_{(s, a_i) \sim \mathcal{D}}D_{KL}(q(z|a_i), p_a(z|s_t))$. To train a robust skill policy, preprocessing steps are applied to the input image, including random cropping, downsizing, and noise addition (Fig. \ref{fig:skill_net}). Subsequently, the skill prior generates skill embedding actions from a concatenated vector comprising image features extracted from ResNet18 \cite{he2016deep} and the robot state.

In test phase, the skill encoder is not utilized. Instead, the skill prior infers a 12-dimensional skill action $z_t=p_a(z|o_t, s_t)$, which is subsequently decoded by the skill decoder into a H-steps (set to 10) robot action trajectory for application on the physical robot. The decoded robot action ($H$ by 9) encompasses the relative positional difference $\{\partial{x_a}, \partial{x_y}, \partial{x_z}\}$ and quaternion rotational difference $\{x_q, x_z, z_q, w_q\}$ of the end-effector, grip action $g_a$ and configuration change action $c_a$ that facilitates the alteration of joint configurations between the forward and downward base poses by a button press on the VR controller. As a result, a single skill action inference leads to the execution of a series of actual robot actions spanning $H$-steps as $p_d(\bold{a_i}|z_t)=\left[ a_t, \dots, a_{t+h}, \dots, a_{t+H} \right]$ and $z_{t+1} \cong z_{t+H}$. The skill encoder is composed of a recurrent layer and two linear layers (256-dim, 128-dim), while the skill prior and skill decoder are constructed using three linear layers (256-dim) with leaky-ReLU \cite{xu2015empirical} activation function.



\subsection{Uncertainty-Aware Conservative Skill Inference}
To address the uncertainties in dynamic environments, we applied Monte-Carlo dropout \cite{gal2016dropout} to the skill prior network of HSN. The skill uncertainty associated with the current observation is determined through the standard deviation of the determinants of K covariance matrices: 

\begin{equation} \label{skill_ucn}
\xi=\text{std}\left( \mathbb{H} \right)
\end{equation}

\noindent where $\mathbb{H}=\{|\Sigma_1|, \dots, |\Sigma_k|, \dots, |\Sigma_{K}| \}$, the K is the number of samples in MC-dropout process and the covariance matrices are derived from the sampled skill actions, which conform to a multivariate Gaussian distribution $z_t \sim \mathcal{N}(\boldsymbol{\mu}, \boldsymbol{\Sigma})$. We then normalized the inferred skill uncertainty to a value between 0 and 1 as follows:
\begin{equation} \label{skill_ucn_normalize}
\hat{\xi} = 1 - \text{exp}(-\epsilon \: \xi)
\end{equation}
\noindent $\epsilon$ is a tunable constant parameter (set to 2e-3). The normalized skill uncertainty is subsequently employed to modulate the level of conservatism in skill planning, encompassing both skill actions and robot actions as 
\begin{align} \label{conservative_skill_inference}
\hat{z_t} &= (1 - \hat{\xi}) \: z_t + \hat{\xi} \: z_{t-1} \\
\hat{a}_{t+h} &= \left( \frac{1}{1+\hat{\xi}} \right) \: a_{t+h}
\end{align}
In the skill embedding space, the conservative skill inference enables a more deliberate skill planning by deducing skill actions that depend on the preceding skill when uncertainty is high. Similarly, according to the last equation, reducing action execution speed by up to 50\% of the maximum can enhance robot operation stability in uncertain situations. The detailed process of uncertainty-aware shared autonomy process, including conservative skill inference, is described in Algorithm 2.

\subsection{Practical Challenges}
A direct numeric “on/off” ablation of the uncertainty term would require running the robot without the conservative gating, exposing it to high‑energy collisions that violate our institute’s safety policy and dramatically slow experimentation (each hard collision triggers inspection and recalibration). In addition, uncertainty manifests only in out‑of‑distribution states, which are by definition hard to reproduce deterministically. For these reasons we evaluate the policy in an ecological setting—dynamic object motions that naturally create distributional shift—and measure downstream stability. This methodology reflects standard practice in safety‑critical robotics and, as shown later, provides strong qualitative evidence of the benefit.

\begin{table}[]
\centering
\caption{Position and orientation constraints in each configuration mode (units: meters and degrees)}
\label{table:const_min_max_thres_all}
\begin{tabular}{cl||l|l|l|l|l|l}
\hline
\multicolumn{2}{c|}{Configuration}              & $x$    & $y$    & $z$    & $\alpha$    & $\beta$   & $\gamma$   \\ \hline
\multicolumn{1}{c|}{\multirow{2}{*}{Fwd}} & min & 0.38 & -0.2 & 0.07 & -135 & -5  & -45 \\ \cline{2-8} 
\multicolumn{1}{c|}{}                     & max & 0.53 & \ 0.2  & 0.3 & \ 135  & \ 20  & \ 45  \\ \hline
\multicolumn{1}{c|}{\multirow{2}{*}{Dwd}} & min & 0.2  & -0.2 & 0.04 & -20  & -40 & -90 \\ \cline{2-8} 
\multicolumn{1}{c|}{}                     & max & 0.44 & \ 0.2  & 0.13 & \ 20   & \ 3   & \ 90  \\ \hline
\end{tabular}
\end{table}


\setlength{\textfloatsep}{5pt}
\begin{algorithm}
    \setstretch{0.9}
    \DontPrintSemicolon 
    \text{Load pretrained networks} $p_d \: \text{and} \: \pi_{N_1} \gets p_a$ \;
    $\mathcal{D} \gets \mathcal{D}_{BC}, \: \mathbb{H} \gets [] $ \;
    \For{\text{epoch} $i=1:L$} {
        \For{\text{rollout} $j=1:M$} {
            \For{\text{timestep} $t\in T \: \text{of rollout} \: j$ } {
                \If{\text{expert has control}}{
                    $\mathcal{D}_j \gets \pi_{E}(x)$
                }
                \Else{
                    $\mathbb{H} \gets []$ \;
                    \For{\text{drops} $k=1:K$} {
                        $z_k=p_{a}(o_t, s_t) \sim \mathcal{N}(\boldsymbol{\mu}, \boldsymbol{\Sigma})$ \;
                        $\text{append} |\boldsymbol{\Sigma}_k| \: \text{of} \: z_k \: \text{to} \: \mathbb{H}$
                    }
                    $\hat{\xi} \gets 1-\text{exp}(-\epsilon \: \text{std}(\mathbb{H}))$ \;
                    $\hat{z_t} \gets (1-\hat{\xi}) z_t + \hat{\xi} \: z_{t-1}$ \;
                    $\hat{a}_{t+h} \gets (1/(1+\hat{\xi})) a_{t+h}$ \;
                }
            }
            $\mathcal{D} \gets \mathcal{D} \cup \mathcal{D}_j$ \;
        }
        $\pi_{N_{i+1}} \gets \text{update} \: \pi_{N_{i}}$ \;
    }
    \caption{Uncertainty-aware Shared Autonomy Process}
\end{algorithm}

\section{EXPERIMENTS}

\subsection{Experimental Setup}
In our setup (Fig. \ref{fig:architecture}), we used a UR3 robot with a Robotiq 2F-85 gripper for manipulation, an HTC VIVE Pro2 for teleoperation, and a RealSense D435 camera for observation. We adjusted the TCP origin by adding a 127mm offset to its z-axis. Two cameras were placed, one in front for recording evaluation videos and one on the rear side (Fig. \ref{fig:skill_net}). Teleoperation and skill learning were conducted on a desktop system with an i7-Xeon processor and an RTX-3090 GPU. For the pick-and-place task, we utilized a white basket with toy fruit, while the pouring task involved a green plastic cup and a transparent bottle containing red beads.

\subsection{Rotation Constraints Validation}
\begin{figure}
	\begin{center}
		\includegraphics[width=\linewidth]{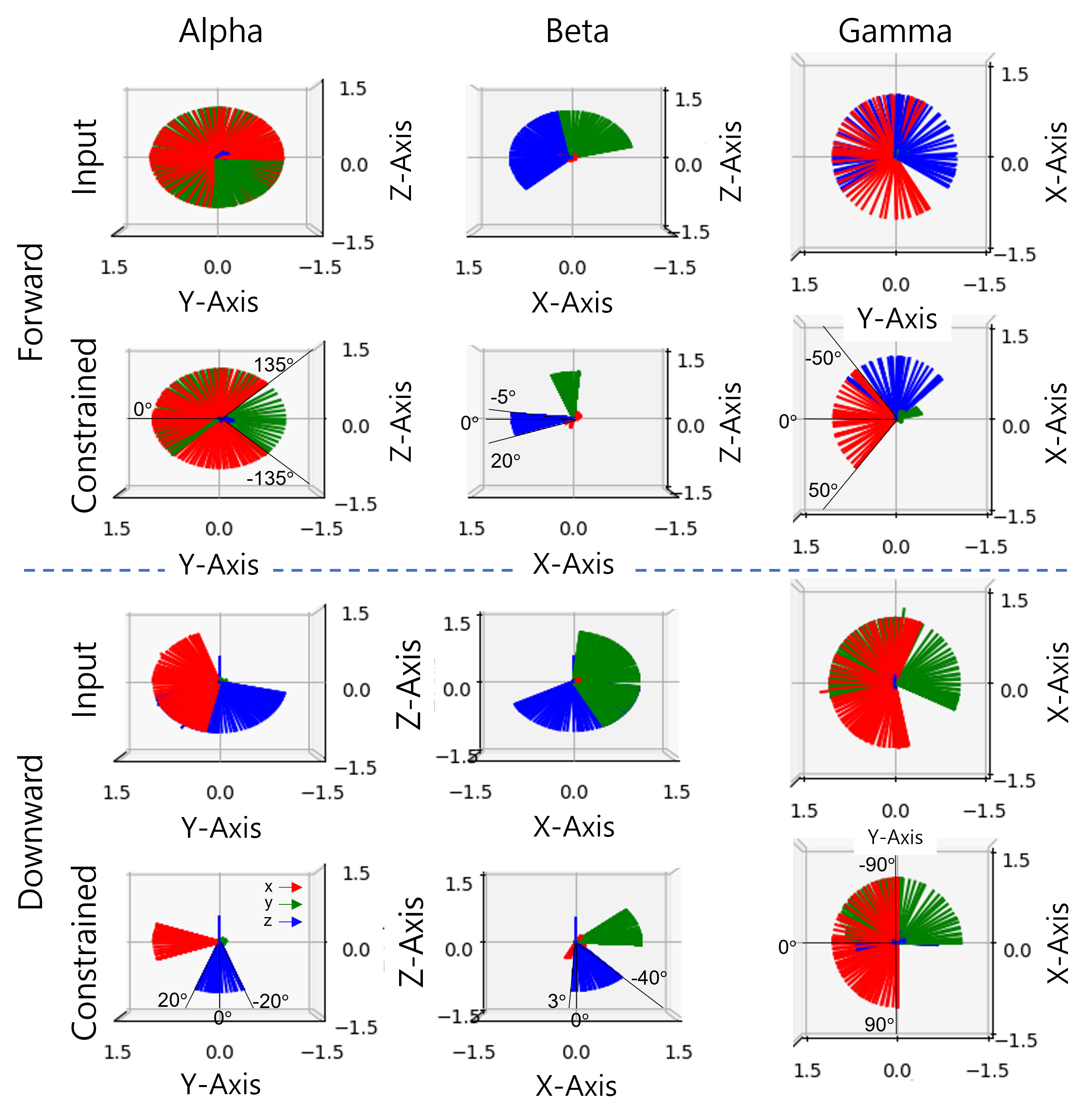}
	\end{center}
	\vspace{-0.5em}
	\caption{Accumulated plot illustrating the input and constrained TCP coordinates for alpha, beta, and gamma orientations.}
        \label{fig:rot_const}
	\vspace{-0.0em}
\end{figure}

Since the position constraints yielded obvious results, we only conducted an evaluation of the orientation constraints. Fig. \ref{fig:rot_const} illustrates the comparative results between the original and constrained rotation motions of the TCP along each orientation axis. The results demonstrate that the proposed rotation constraint algorithm effectively confines the input rotations to the specified limits as detailed in Table \ref{table:const_min_max_thres_all}.



\begin{figure*}
        \centering
        \includegraphics[width=1.0\linewidth]{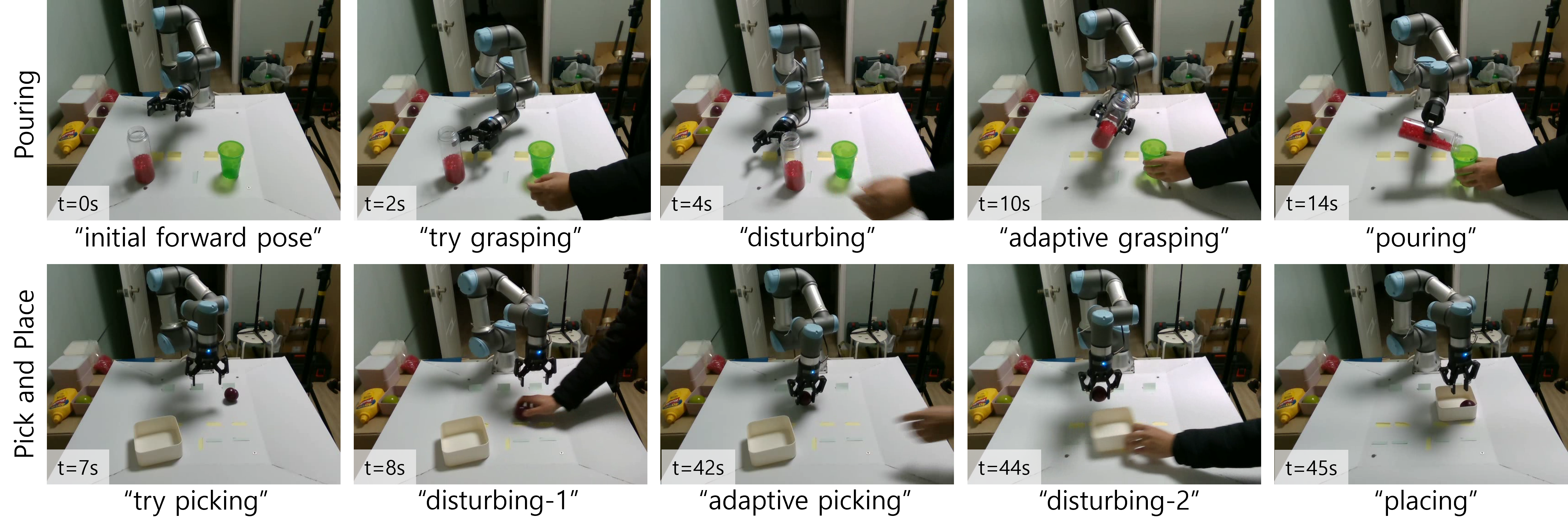}
	\vspace{-1.5em}
	\caption{Evaluation of adaptable manipulation skills in dynamic environments where humans manipulate target objects to simulate variations.}
        \vspace{-1.0em}
        \label{fig:task_eval_dynamic}
\end{figure*}

\subsection{Learning Pouring and Pick and Place Skills}
We conducted HSN training over 10K epochs using 308 and 283 demonstration datasets for pouring and pick-and-place tasks. Each demonstration began with randomly positioned target objects within predefined manipulable regions (Fig. \ref{fig:skill_net}) for respective tasks. The initial robot configuration was also randomized (forward or downward) with uniform noise. The first half of the dataset was collected via VR teleoperation for initial skill policy training. In the remaining half, correction motion datasets were gathered by user interventions during SAP-based task execution in situations where collisions and task failures (e.g., tipping over a water bottle) were expected. We independently trained pouring and pick-and-place skills and achieved 90\% and 80\% success rates, respectively, in 10 disturbance-free trials. Failures resulted from minor spatial errors despite correct semantic actions (e.g., approaching the bottle correctly, but failed due to a few spatial errors). This indicates success in semantic learning, with spatial errors expected to decrease with a more extensive and varied dataset. For pick-and-place, occlusion caused by robot hardware made tasks challenging, leading to lower performance compared to pouring, which is relatively simpler. To tackle this, we plan to use multiple cameras for observation in future work.




\subsection{Task Performance in Dynamic Environment}
We assessed our skills in dynamic environments, where target objects were moved during tasks. Despite deliberate disruptions, the agent achieved success rates of 80\% and 70\% (Fig. \ref{fig:task_eval_dynamic}, Table \ref{table:exp_result}). Our HSN demonstrated the ability to adapt to dynamic changes, even without specific disruption demonstrations in the initial dataset. However, adapting to pose variations, like recovering a fallen bottle, would require additional demonstrations.

\subsection{Multi-Skill Learning and Task Transition}
We evaluated the HSN's ability to learn multi-configurable skills. Initially, we trained it for 5K epochs using demonstration datasets from two tasks. Then, we collected 200 additional demonstrations with SAP, totaling 10K epochs of training. The results demonstrated the HSN's success in performing pouring and pick-and-place tasks, achieving 80\% and 90\% success rates, respectively, over 10 trials. The agent smoothly transitioned configurations between tasks within a few seconds, achieving a 100\% success rate in configuration transitions. The outcomes from both single and multi-skill learning, detailed in Table \ref{table:exp_result}, suggest the potential extension of this method to a broader range of skills and objects.

\subsection{Qualitative stability study}
To probe the effect of Conservative Skill Inference (CSI) without compromising hardware safety, we trained two policies on the same 70‑trajectory core dataset: CSI‑ON (ours) and CSI‑OFF (dropout disabled, no speed scaling). After 1k epochs we executed 30 validation roll‑outs.
Observations. CSI‑OFF diverged in 9 cases, causing either self‑collision or abrupt joint‑limit hits that forced an emergency stop. CSI‑ON finished all roll‑outs; in high‑uncertainty moments it slowed the end‑effector by 40\% (median), giving the human operator ample time to intervene. The accompanying video highlights the contrast.
Take‑away. Even without a large‑scale numeric ablation, the qualitative gap is stark and aligns with our design hypothesis: latent uncertainty can be transformed into actionable caution at run‑time









\begin{table}[]
\centering
\caption{Evaluation results for pouring, pick-and-place tasks, and the multi-skill agent in both static and dynamic settings, along with task transition success rates (Unit: \%)
}
\label{table:exp_result}
\begin{tabular}{c|lll}
\hline
\multirow{2}{*}{Task} & \multicolumn{3}{c}{Condition}        \\ \cline{2-4} 
                      & \multicolumn{1}{l|}{Static} & \multicolumn{1}{l|}{Dynamic} & Task Transition \\ \hline
Pouring               & \multicolumn{1}{l|}{90}     & \multicolumn{1}{l|}{80}      & - \\ \hline
Pick and place         & \multicolumn{1}{l|}{90}     & \multicolumn{1}{l|}{70}     & -  \\ \hline
Multi-Skill         & \multicolumn{1}{l|}{80}     & \multicolumn{1}{l|}{70}     & 100  \\ \hline
\end{tabular}
\end{table}

\section{DISCUSSION and CONCLUSIONS}
\noindent \textit{Ablation‑free justification.} The conservative gating is intrinsically a safety feature; turning it off solely for a clean ablation would defeat its purpose and violate lab safety protocols. Instead, we rely on qualitative evidence (Section 4.4) and task‑level metrics. A simulation‑only replica, where destructive failures are cheap, is promising future work.

In this paper, we propose a learning method within the shared autonomy process, where skills are acquired through human demonstration and correction. This method is based on the uncertainty of manipulation skills, enabling conservative task execution to expand the permissible margin for human errors. Additionally, we implement a shared autonomy system for robot manipulation skill learning, a key component that has shown recent outstanding results but lacks specific public details. Through our proposed system, we experimentally demonstrate the learning of multi-configurable manipulation skills and the ability to perform skill replanning for task completion in dynamic environments with disturbances.

We introduce a hierarchical skill network to infer uncertainty in the current context at an abstract level. We also propose a technique for conservative skill inference using MC-dropout based uncertainty estimation for skill layers and terminal output actions. This approach provides more flexibility in assessing the timing of human intervention and mitigates the potential for errors, out-of-distribution scenarios, and risk-related task failures.

The proposed system enables more stable manipulation skill learning through conservative skill inference. However, because it utilizes features from the entire video as observations, there are instances where uncertainty inference for partial changes becomes uncertain itself. In future research, we plan to incorporate video understanding methods at the patch level, such as Vision Transformer \cite{dosovitskiy2020image}, to enhance the performance of skill uncertainty.

\addtolength{\textheight}{-12cm}   





\section*{ACKNOWLEDGMENT}

This work was supported by Institute of Information \& communications Technology Planning \& Evaluation(IITP) grant funded by the Korea government(MSIT) (No.2020-0-00842, Development of Cloud Robot Intelligence for Continual Adaptation to User Reactions in Real Service Environments, 50\%) and (No. 2022-0-00951, Development of Uncertainty-Aware Agents Learning by Asking Questions, 50\%)

\bibliographystyle{IEEEtran}
\bibliography{refs.bib}

\begin{thebibliography}{10}
\providecommand{\url}[1]{#1}
\csname url@rmstyle\endcsname
\providecommand{\newblock}{\relax}
\providecommand{\bibinfo}[2]{#2}
\providecommand\BIBentrySTDinterwordspacing{\spaceskip=0pt\relax}
\providecommand\BIBentryALTinterwordstretchfactor{4}
\providecommand\BIBentryALTinterwordspacing{\spaceskip=\fontdimen2\font plus
\BIBentryALTinterwordstretchfactor\fontdimen3\font minus \fontdimen4\font\relax}
\providecommand\BIBforeignlanguage[2]{{%
\expandafter\ifx\csname l@#1\endcsname\relax
\typeout{** WARNING: IEEEtran.bst: No hyphenation pattern has been}%
\typeout{** loaded for the language `#1'. Using the pattern for}%
\typeout{** the default language instead.}%
\else
\language=\csname l@#1\endcsname
\fi
#2}}

\bibitem{qureshi2023imitation}
O.~Qureshi, M.~N. Durrani, and S.~A. Raza, ``Imitation learning for autonomous driving cars,'' in \emph{2023 3rd International Conference on Artificial Intelligence (ICAI)}.\hskip 1em plus 0.5em minus 0.4em\relax IEEE, 2023, pp. 58--63.

\bibitem{wang2022adaptive}
Y.~Wang, C.~C. Beltran-Hernandez, W.~Wan, and K.~Harada, ``An adaptive imitation learning framework for robotic complex contact-rich insertion tasks,'' \emph{Frontiers in Robotics and AI}, vol.~8, p. 777363, 2022.

\bibitem{stepputtis2020language}
S.~Stepputtis, J.~Campbell, M.~Phielipp, S.~Lee, C.~Baral, and H.~Ben~Amor, ``Language-conditioned imitation learning for robot manipulation tasks,'' \emph{Advances in Neural Information Processing Systems}, vol.~33, pp. 13\,139--13\,150, 2020.

\bibitem{jang2022bc}
E.~Jang, A.~Irpan, M.~Khansari, D.~Kappler, F.~Ebert, C.~Lynch, S.~Levine, and C.~Finn, ``Bc-z: Zero-shot task generalization with robotic imitation learning,'' in \emph{Conference on Robot Learning}.\hskip 1em plus 0.5em minus 0.4em\relax PMLR, 2022, pp. 991--1002.

\bibitem{kelly2019hg}
M.~Kelly, C.~Sidrane, K.~Driggs-Campbell, and M.~J. Kochenderfer, ``Hg-dagger: Interactive imitation learning with human experts,'' in \emph{2019 International Conference on Robotics and Automation (ICRA)}.\hskip 1em plus 0.5em minus 0.4em\relax IEEE, 2019, pp. 8077--8083.

\bibitem{zhao2023survey}
W.~X. Zhao, K.~Zhou, J.~Li, T.~Tang, X.~Wang, Y.~Hou, Y.~Min, B.~Zhang, J.~Zhang, Z.~Dong, \emph{et~al.}, ``A survey of large language models,'' \emph{arXiv preprint arXiv:2303.18223}, 2023.

\bibitem{brohan2022rt}
A.~Brohan, N.~Brown, J.~Carbajal, Y.~Chebotar, J.~Dabis, C.~Finn, K.~Gopalakrishnan, K.~Hausman, A.~Herzog, J.~Hsu, \emph{et~al.}, ``Rt-1: Robotics transformer for real-world control at scale,'' \emph{arXiv preprint arXiv:2212.06817}, 2022.

\bibitem{brohan2023rt}
A.~Brohan, N.~Brown, J.~Carbajal, Y.~Chebotar, X.~Chen, K.~Choromanski, T.~Ding, D.~Driess, A.~Dubey, C.~Finn, \emph{et~al.}, ``Rt-2: Vision-language-action models transfer web knowledge to robotic control,'' \emph{arXiv preprint arXiv:2307.15818}, 2023.

\bibitem{bradbury2016attention}
N.~A. Bradbury, ``Attention span during lectures: 8 seconds, 10 minutes, or more?'' 2016.

\bibitem{reason2000human}
J.~Reason, ``Human error: models and management,'' \emph{Bmj}, vol. 320, no. 7237, pp. 768--770, 2000.

\bibitem{pertsch2021accelerating}
K.~Pertsch, Y.~Lee, and J.~Lim, ``Accelerating reinforcement learning with learned skill priors,'' in \emph{Conference on robot learning}.\hskip 1em plus 0.5em minus 0.4em\relax PMLR, 2021, pp. 188--204.

\bibitem{gal2016dropout}
Y.~Gal and Z.~Ghahramani, ``Dropout as a bayesian approximation: Representing model uncertainty in deep learning,'' in \emph{international conference on machine learning}.\hskip 1em plus 0.5em minus 0.4em\relax PMLR, 2016, pp. 1050--1059.

\bibitem{csiszar1975divergence}
I.~Csisz{\'a}r, ``I-divergence geometry of probability distributions and minimization problems,'' \emph{The annals of probability}, pp. 146--158, 1975.

\bibitem{he2016deep}
K.~He, X.~Zhang, S.~Ren, and J.~Sun, ``Deep residual learning for image recognition,'' in \emph{Proceedings of the IEEE conference on computer vision and pattern recognition}, 2016, pp. 770--778.

\bibitem{xu2015empirical}
B.~Xu, N.~Wang, T.~Chen, and M.~Li, ``Empirical evaluation of rectified activations in convolutional network,'' \emph{arXiv preprint arXiv:1505.00853}, 2015.

\bibitem{dosovitskiy2020image}
A.~Dosovitskiy, L.~Beyer, A.~Kolesnikov, D.~Weissenborn, X.~Zhai, T.~Unterthiner, M.~Dehghani, M.~Minderer, G.~Heigold, S.~Gelly, \emph{et~al.}, ``An image is worth 16x16 words: Transformers for image recognition at scale,'' \emph{arXiv preprint arXiv:2010.11929}, 2020.

\end{thebibliography}

\end{document}